%% file: main.tex
\documentclass[letterpaper]{article} % DO NOT CHANGE THIS
\usepackage{aaai25}  % DO NOT CHANGE THIS
\usepackage{times}  % DO NOT CHANGE THIS
\usepackage{helvet}  % DO NOT CHANGE THIS
\usepackage{courier}  % DO NOT CHANGE THIS
\usepackage[hyphens]{url}  % DO NOT CHANGE THIS
\usepackage{graphicx} % DO NOT CHANGE THIS
\usepackage{natbib}  % DO NOT CHANGE THIS AND DO NOT ADD ANY OPTIONS TO IT
\usepackage{caption} % DO NOT CHANGE THIS AND DO NOT ADD ANY OPTIONS TO IT
\usepackage{algorithm}
\usepackage{algorithmic}
\usepackage{amsmath}
\usepackage{helvet}
\usepackage[latin9]{inputenc}
\usepackage{array}
\usepackage{booktabs}
\usepackage{multirow}
\usepackage{amstext}

\usepackage{newfloat}
\usepackage{listings}
\DeclareCaptionStyle{ruled}{labelfont=normalfont,labelsep=colon,strut=off} % DO NOT CHANGE THIS
\floatstyle{ruled}
\newfloat{listing}{tb}{lst}{}
\floatname{listing}{Listing}
\usepackage{bibentry}
\makeatother

\begin{document}
\title{Progressive Multi-granular Alignments for Grounded Reasoning in Large Vision-Language Models}
\author {
    % Authors
    Quang-Hung Le\textsuperscript{\rm 1},
    Long Hoang Dang\textsuperscript{\rm 2},
    Ngan Le\textsuperscript{\rm 3},
    Truyen Tran\textsuperscript{\rm 1},
    Thao Minh Le\textsuperscript{\rm 1}
}
\affiliations{
    % Affiliations
    \textsuperscript{\rm 1}Applied Artificial Intelligence Institute (A$^2$I$^2$), Deakin University, Australia.\\
    \textsuperscript{\rm 2}Posts and Telecommunications Institute of Technology, Vietnam.\\
    \textsuperscript{\rm 3}University of Arkansas, USA.\\
    q.le@deakin.edu.au, longdh@ptit.edu.vn, thile@@uark.edu, truyen.tran@deakin.edu.au, thao.le@deakin.edu.au
}

\maketitle
\global\long\def\TaskName{\text{LVLMs}}%
\global\long\def\Dataset{\text{CompoVL}}%
\global\long\def\ModelName{\text{PromViL}}%
\global\long\def\TestData{\text{CompoVL-hard}}%

\begin{abstract}
\input{abstract.tex}

\end{abstract}

\section{Introduction}

\input{intro.tex}

\section{Related Works}

\input{related_work.tex}

\section{Preliminaries}

\input{prelim.tex}

\section{Methods}

\input{method.tex}

\section{Experiments}

\input{experiment.tex}

\section{Conclusion}

\input{discussion.tex}

\bibliography{aaai25}

% \section*{Reproducibility Checklist}

% \input{reproducibility_checklist.tex}
\end{document}

%% file: abstract.tex
Existing Large Vision-Language Models (LVLMs) excel at matching concepts
across multi-modal inputs but struggle with compositional concepts
and high-level relationships between entities. This paper introduces
Progressive multi-granular Vision-Language alignments ($\ModelName$),
a novel framework to enhance LVLMs' ability in performing grounded
compositional visual reasoning tasks. Our approach constructs a hierarchical
structure of multi-modal alignments, ranging from simple to complex
concepts. By progressively aligning textual descriptions with corresponding
visual regions, our model learns to leverage contextual information
from lower levels to inform higher-level reasoning. To facilitate
this learning process, we introduce a data generation process that
creates a novel dataset derived from Visual Genome, providing a wide
range of nested compositional vision-language pairs. Experimental
results demonstrate that our PromViL framework significantly outperforms
baselines on various visual grounding and compositional question answering
tasks. The code is available at: https://github.com/lqh52/PromViL.

%% file: intro.tex
Large Vision-Language Models (LVLMs) \cite{liu2024visual,li2023blip},
pre-trained on vast amounts of image-text data, hold great
promise in solving complex vision-language (V-L) tasks. However, current
LVLMs still fall short in compositional reasoning -- the ability
to answer complex queries composed of smaller elements \cite{ma2023crepe,zhao2022vl}.
Compositionality is a pervasive phenomenon, seen in language with
syntax trees, in vision with scenes and objects, and indeed wherever
``the meaning of the whole is a function of the meanings of its parts''
\cite{cresswell2016logics}.

When faced with a complex visual query, humans can recursively decompose
it into smaller components while simultaneously grounding them to
corresponding visual elements \cite{bottou2014machine,hupkes2020compositionality,janssen1997compositionality}.
Current LVLMs do not yet possess that ability. Most existing
models \cite{huang2024language,li2023blip} process whole textual
prompts and entire images, relying on associations between holistic
image embeddings and textual embeddings. These methods disregard
interactions between sentence parts and image components, leading
to poor grounding of textual information \cite{li2023covlm}
and subsequently to suboptimal performance in reasoning tasks.

\begin{figure}[t]
\begin{centering}
\resizebox{!}{0.8\columnwidth}{\includegraphics[width=1\columnwidth]{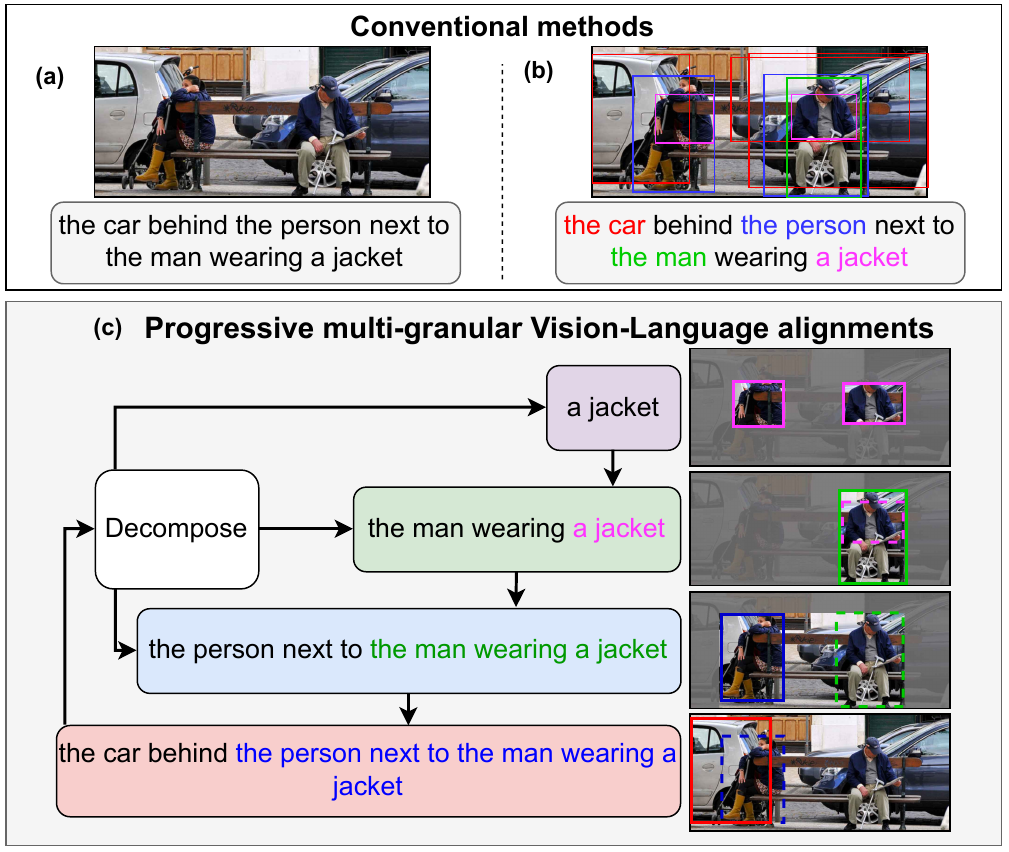}}
\par\end{centering}
\centering{}\caption{\textbf{Comparison with Existing LVLMs:} (a) Coarse-grained: Whole
image/region with full text, lacks object details. (b) Fine-grained:
Simple phrases and bounding boxes, lacks relational context. (c) $\protect\ModelName$
employs hierarchical multi-granular associations, progressively utilizing
simpler concepts as cues to understand more complex ones.\label{fig:teaser}}
\end{figure}

Others, like Kosmos-2 \cite{peng2024grounding}, Pink \cite{xuan2024pink},
and CoVLM \cite{li2023covlm} address these drawbacks by infusing
location information corresponding to visual entities into the language
generation process to enhance grounding ability. As a result, they
can ground simple concepts involving individual objects, but still
struggle with scenarios involving multiple objects and complex relationships.
This limitation may stem from their underlying grounding processes.
The coarse-grained processes (Fig. \ref{fig:teaser}.a) align complex
expressions (e.g., "the car behind the person next to
the man with a jacket") to a region but ignore object-level
alignments, which is crucial for tasks like visual reasoning and image
captioning \cite{zeng2021multi}. The fine-grained processes (Fig.
\ref{fig:teaser}.b), on the other hand, align concepts with single
objects but struggle to understand relations among multiple objects,
especially in ambiguous scenarios (e.g., identifying ``man with a
jacket'' when multiple men are present). Both approaches inadequately
capture nuanced relationships between textual descriptions and visual
elements, limiting effectiveness in tasks requiring sophisticated
compositional reasoning.

Addressing these limitations, we propose a novel compositional reasoning
framework that leverages off-the-shelf LVLMs in an `agentic flow'
style. Our approach, applicable to any grounded LVLM, integrates \textit{multi-granular
language-vision training} with \textit{progressive reasoning}: \textgreater We
consolidate the model's grounding ability through a dataset construction
pipeline that leverages the Visual Genome dataset \cite{krishna2017visual}
to curate nested compositional V-L pairs. This enables training on
multiple complexity levels, allowing models to learn visual relations
of unbounded complexity. Unlike previous methods, it is not constrained
by the number of concepts in textual information or limited to specific
levels of visual information. \textgreater Rather than inputting
the entire question and image at once, we decompose the text into
object-centric components with gradually increasing complexity. We
then prompt the model to perform step-by-step reasoning, progressing
from simple to complex. Visual information from each step is fed back
to the model in subsequent steps, gradually guiding it through grounding
and reasoning processes. We call the method $\ModelName$ (\textbf{Pro}gressive
\textbf{m}ulti-granular \textbf{Vi}sion-\textbf{L}anguage alignments)
and show its behaviors in Fig.~\ref{fig:teaser}.

We conduct extensive experiments to demonstrate the effectiveness
of our framework: Compared to other models of the same size, $\ModelName$,
with only 4.9\% tunable parameters and 60K fine-tuning data samples,
shows significant improvements: an approximately 9.0 point increase
on our benchmark; up to 5.5 on zero-shot grounding tasks, respectively;
and nearly 5 point and 10 point increases in accuracy and validity,
respectively, on zero-shot compositional reasoning task. Importantly,
$\ModelName$ surpasses larger models like CoVLM (2.4B) and Pink (7B)
on grounding tasks, and exceeds baselines fine-tuned with twice the
VQA data on compositional reasoning tasks.

Our main contributions are threefold: (1) We propose $\ModelName$,
a novel framework integrating multi-granular language-vision training
with progressive reasoning, allowing models to ground and reason in
scenarios with intricate textual information and multiple visual relations.
(2) We introduce a dataset construction pipeline to create a new dataset
of nested compositional V-L pairs curated from Visual Genome, enabling
training on multiple complexity levels. (3) We conduct extensive experiments
demonstrating $\ModelName$'s effectiveness in handling complex visual
scenes and linguistic descriptions, outperforming existing approaches
on various benchmarks. Our experiments are reproducible in academia,
using only public data and models. The model can be trained on consumer
GPUs with 32GB memory. We will release our code and datasets to support
further research in this field.

%% file: related_work.tex
\textbf{Large Vision-Language Models} (LVLMs) have shown impressive
capabilities in tasks like image captioning and visual question answering
\cite{liu2024visual,bai2023qwen,wang2022ofa}. Recent methods enable
phrase-region alignments, allowing models to perceive image regions
and ground text to visual entities \cite{chen2023minigpt,li2023covlm,peng2024grounding,xuan2024pink}.
However, these approaches often struggle with compositional tasks
involving concepts of varying complexity attended by multiple objects
and relations \cite{yuksekgonul2023and,li2023covlm}. Our method addresses
this limitation by introducing hierarchical multi-granular V-L alignments
and progressive reasoning, enabling multi-step grounding and reasoning
from simple to complex tasks. Previous studies have shown the benefits
of utilizing multi-granularity information in training LVLMs in improving
textual and visual alignments \cite{zeng2021multi,le2020hierarchical,dang2021hierarchical,gao2022pyramidclip,chen2024lion}.
However, existing approaches often use single granularity levels or
obtain information in embedding space, contradicting the varying structure
of text and images. Our method derives granularity levels based on
language structure, creating a suitabletailored hierarchical representation
for each input. This enables the model to learn multimodal alignments
across granularities, enhancing its ability to reason across textual
and visual inputs effectively.

\textbf{Grounding datasets used in LVLMs}  can be broadly categorized
into two groups: \textgreater The coarse-grained group such as RefCOCO/+
\cite{kazemzadeh2014referitgame}, RefCOCOg \cite{mao2016generation},
and GRIT \cite{peng2024grounding}, provide pairs of phrases and corresponding
image bounding boxes. However, for phrases containing multiple objects,
they lack bounding boxes for individual elements. As a result, models
trained on these datasets struggle to learn fine-grained alignments
between vision and language, e.g., object level \cite{zeng2021multi}.
\textgreater The fine-grained group such as Flickr30K \cite{plummer2015flickr30k}
and Objects365 \cite{shao2019objects365} provide simple phrases describing
single objects (e.g., ``the person'', ``the jacket'') with corresponding
bounding boxes. While these datasets enable object-centric feature
learning, they often fail to represent relationships among multiple
objects effectively (e.g., ``the person with jacket''). This becomes
particularly problematic when dealing with ambiguous images that require
relational context to locate correct entities (e.g., multiple ``persons'',
as illustrated in Fig.~1).

Several LVLMs have utilized both groups of datasets \cite{chen2023minigpt,xuan2024pink},
but they still fall short as each sample maintains a single granularity
level, and the fine-grained and coarse-grained data remain unrelated
to each other within samples. Our data generation approach addresses
this limitation, resulting in a new dataset named CompoVL, based on
the Visual Genome dataset \cite{krishna2017visual}. CompoVL provides
multi-grained data for each data instance, offering richer information
for training and aligning well with our method's objectives.

%% file: prelim.tex
\begin{figure*}
\begin{centering}
\resizebox{0.9\textwidth}{!}{\includegraphics[width=1\textwidth]{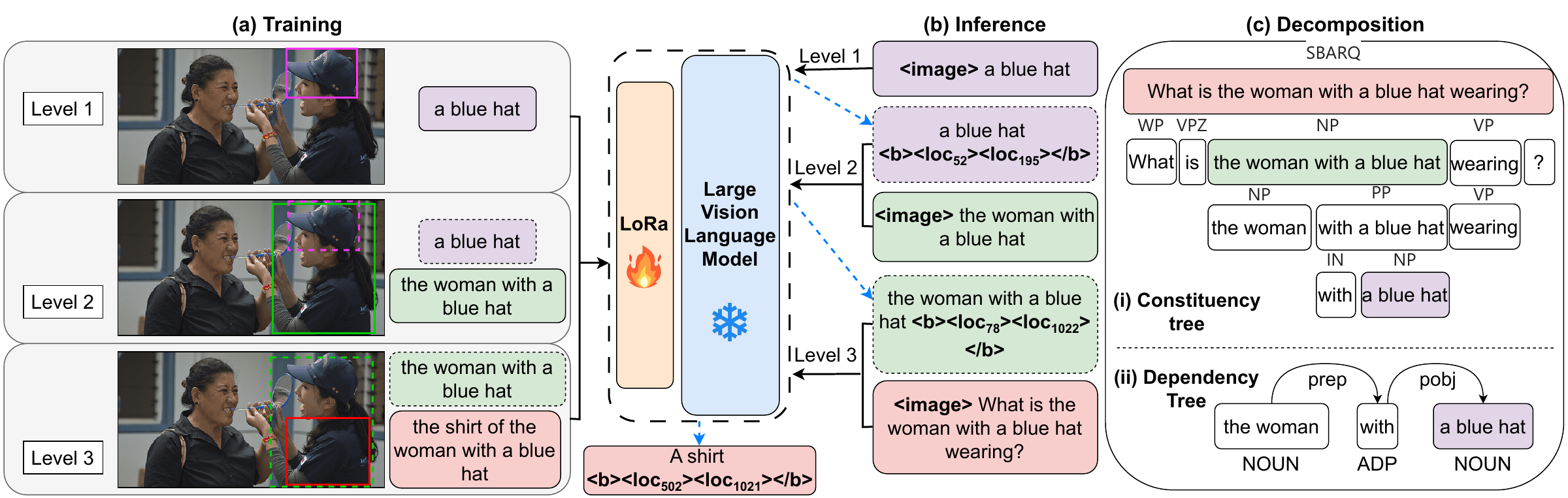}}
\par\end{centering}
\centering{}\caption{\label{fig:Our-method}Overview of our $\protect\ModelName$ framework.
(a) Training: Learn multi-level visual entities-textual expression
associations. (b) Inference: Progressively prompt from simple to complex,
using prior responses as clues. (c) Decomposition: Extract nested
subsequences based on (i) constituency parsing (simplified for illustration)
and (ii) dependency parsing.}
\end{figure*}

Most state-of-the-art $\TaskName$ such as LlaVa \cite{liu2024visual},
Flamingo \cite{alayrac2022flamingo}, and BLIP-2 \cite{li2023blip},
are extensions of language-based LLMs designed to generate text responses
to input images and text prompts. While capable of handling diverse
vision-to-language tasks, these models only scratch the surface of
holistic visual scene understanding and fall short in fine-grained
comprehension of specific visual regions of interest. 

A new family of $\TaskName$ has emerged to address these drawbacks
by producing visual answers such as bounding boxes or segmentation
masks. Examples include Kosmos-2 \cite{peng2024grounding}, Pink \cite{xuan2024pink},
and MiniGPTv2 \cite{chen2023minigpt}. By locating specific image
regions, these models alleviate ambiguities of text-only descriptions
and offer insights into the decision-making processes of $\TaskName$. 

To achieve this, these models process multimodal token sequences combining
text spans and their corresponding spatial locations (e.g., bounding
box coordinates), which are placed next to each other. Trained on
vast grounded image-text datasets using next-token prediction, they
are able to generate both text descriptions of the region of interest
and spatial tokens indicating object location within the image. For
instance, for the query "The woman with a blue hat"
and its bounding boxes correspondence, Kosmos-2 employs a markdown-like
input format: \textsl{``\textless s\textgreater\textless img\textgreater Image
Embedding\textless /img\textgreater{} \textless grounding\textgreater{}
\textless p\textgreater The woman\textless /p\textgreater{} \textless b\textgreater\textless loc}\textsubscript{78}\textsl{\textgreater\textless loc}\textsubscript{1022}\textsl{\textgreater\textless /b\textgreater{}
with \textless p\textgreater a blue hat\textless /p\textgreater}\textbf{\textsl{
}}\textsl{\textless b\textgreater\textless loc}\textsubscript{52}\textsl{\textgreater\textless loc}\textsubscript{195}\textsl{\textgreater\textless /b\textgreater{}
\textless /s\textgreater ''}. Here, \emph{\textless s\textgreater}
and \emph{\textless /s\textgreater} are start and end tokens of
the text sequence; \emph{\textless img\textgreater} and \emph{\textless /img\textgreater}
refer to image embedding; \emph{\textless p\textgreater} and \emph{\textless /p\textgreater}
indicate the boundaries of non-relational concepts within input text
sequence, and \emph{\textless b\textgreater} and\emph{ \textless /b\textgreater}
refer to their corresponding bounding box locations. These models'
output include both text tokens describing the visual content of the
region of interest and spatial location tokens locating the object's
position within the visual scene.

\textbf{Compositional visual reasoning} requires understanding two-way
interactions between sentence parts and corresponding image regions.
However, current $\TaskName$ focus on the alignments of either overly
fine-grained or coarse-grained concepts and image regions. Fine-grained
concepts include non-relational individual concepts, such as ``a
woman'' or ``a blue hat'', while high-order relational concepts
might be ``the shirt of the woman with a blue hat''. This limitation
partly stems from a lack of intermediate connections in existing datasets.
We propose a novel mechanism to obtain such connections, bridging
the generalization gap from fine-grained concepts to coarse-grained
concepts.

%% file: method.tex
\subsection{Progressive multi-granular V-L Alignments}

Given a compositional input sentence, we decompose it into a series
of \emph{nested subsequences}. This input sentence and elements in
the nested subsequences are later referred to as ``expressions''
for the remainder of this paper. These expressions cover concepts
of varying complexity, ranging from individual concepts (e.g., ``a
woman'', ``a blue hat'') to high-order relational concepts (e.g.,
``the woman with a blue hat'', ``the shirt of the woman with a
blue hat''). 

Our approach, termed \textbf{\emph{Pro}}\emph{gressive }\textbf{\emph{m}}\emph{ulti-granular
}\textbf{\emph{Vi}}\emph{sion-}\textbf{\emph{L}}\emph{anguage alignments
(PromViL)}, leverages the alignments between vision-language pairs
of increasing levels of complexity to create a \emph{progressive chain
of reasoning steps} for understanding complex compositional expressions.
Assuming we have access to nested vision-language (V-L) pairs, our
task is to direct $\TaskName$ to iteratively leverage feedback from
lower levels to properly align more complex expressions with their
corresponding visual regions (See Fig.~\ref{fig:Our-method}). The
next section details our method for generating nested V-L pairs from
existing data.

\begin{figure}
\centering{}\includegraphics[width=1\columnwidth]{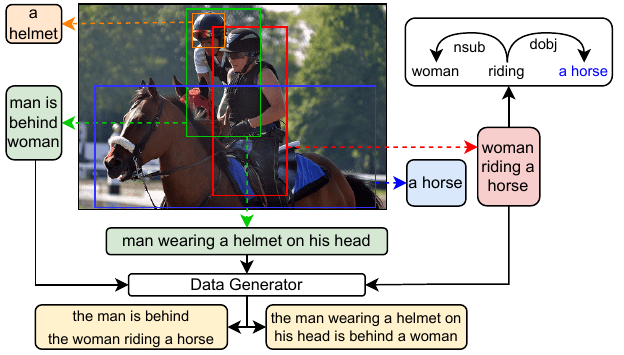}\caption{\label{fig:CompoVG-generation} Multi-granular Compositional V-L Data
Generation. We create a novel dataset with rich, multi-granular V-L
data using existing VG annotations.}
\end{figure}

\subsection{\label{subsec:Compositional-Visual-Genome}Multi-granular Compositional
V-L Dataset }

\textbf{Nested Vision-Language Pairs Generation:} To generate nested
V-L pairs , we utilize Visual Genome (VG) \cite{krishna2017visual}
annotations and an open-source LLM for text generation (Mixtral8x7B
\cite{jiang2024mixtral} in our implementation). Our data generation
pipeline (See Fig.~\ref{fig:CompoVG-generation}) begins by assigning
the level of complexity of an input expressions based on the depth
of relational steps needed to arrive at the main entity of interest.
Level-one expressions include non-relational concepts like "the
woman" or "a horse" , while
level-two expressions combine these with a relationship, such as " the
woman riding a horse" . Our level-one and level-two pairs
use direct VG annotations. For higher levels, we instruct the LLM
to generate text descriptions from VG predicates \emph{\textless subject,
relation, object\textgreater} that share entities in common. For
example, in Fig.~\ref{fig:CompoVG-generation}, to generate the expression
``the man is behind the woman riding a horse'', we first retrieve
predicates \emph{\textless man, behind, woman\textgreater} and
\emph{\textless woman, riding, horse\textgreater}. Then, we instruct
the LLM to produce more complex textual descriptions by combining
and extending information from these two input predicates. To reduce
the effects of hallucinations by LLMs, we direct them to strictly
adhere to the genuine objects and relationships in provided input
predicates and not to invent any new objects and relationships (See
Appendix for our prompts used). To assign corresponding visual bounding
boxes to these generated descriptions, we employ a dependency parser
\cite{nivre-2008-algorithms} to identify the main object of interest
within a given expression. The visual bounding boxes are then chosen
as bounding boxes of the identified main entities in VG. For instance,
the bounding box corresponding to " the man"{}
is chosen as the bounding box for the phrase " the man
is behind the woman riding a horse" . 

From these generated descriptions and VG annotations, we construct
a hierarchical series of nested compositional V-L pairs, progressing
from simple to complex. Lower-level expressions form building blocks
for higher-level ones by extending them as referential components.
This results in a list of nested expressions such as: "a
horse" (level-one), "the woman riding
a horse" (level-two), and "the man is
behind the woman riding a horse" (level-three) (Fig.~\ref{fig:CompoVG-generation}).
Multiple expressions can exist at each level. In total, we generate
29K such lists of nested expressions, comprising up to 115K individual
V-L pairs. \emph{We refer to each list of a compositional V-L pair
and its associated nested subsequences as one data instance in our
dataset}.

To diversify spatial relationships in our dataset beyond VG, we incorporate
annotations from VSR \cite{liu2023visual}. Given VSR's annotations
in the format of \textit{\textless subject, relation, object\textgreater{}
}\textit{\emph{predicates, we }}use GroundingDINO \cite{liu2023grounding}
to obtain bounding boxes for all the involved visual entities, adding
1.2K data instances to our dataset. As $\TaskName$ are often trained
to perform various tasks, we also include 8K and 22K data instances
randomly sampled from VG-VQA and LLaVA-Instruct150K \cite{liu2024visual},
respectively. Our combined dataset, multi-granular Compositional Vision-Language
($\Dataset$), contains a total of 60.3K instances.

To assess the limitations of state-of-the-art $\TaskName$ on compositional
visual grounding, we also provide a $\Dataset$ subset, named \emph{$\TestData$,
}containing only level-two and higher V-L pairs. This comprises 6K
image-expression pairs in total. Given these pairs, $\TaskName$ are
required to output bounding boxes indicating the location of visual
regions corresponding to the given expressions. Compared to RefCOCOg
\cite{mao2016generation}, our dataset has a higher average object
count per linguistic expression (\textbf{2.70} vs. 2.29) and a greater
average complexity level per expression (\textbf{2.56} vs. 2.26),
offering more challenging compositional scenarios. We will later empirically
demonstrate current models' limitations on this subset (See Sec.~\ref{subsec:Experimental-Results}).

\textbf{Annotation's reliability:} To ensure data quality, we randomly
sampled 2\% of $\TestData$ for three independent human evaluations.
Evaluators are asked to answer questions about the generated text
descriptions in terms of: naturalness, ambiguity of visual answers
and bounding box accuracy. The questions asked include: ``Does the
generated image caption sound natural?'', ``Does the caption refer
to a unique object in the image? ``, ``Is the bounding box correct
for the caption?''. According to assessments, 92.5\% of the generated
text descriptions sounds natural while 87.61\% of them refer to unique
visual objects, and 92.48\% have correct bounding boxes. We also measured
inter-annotator agreement using \emph{Kappa scores \cite{cohen1960coefficient}}
and found an average score of \emph{0.68} for question 1, \emph{0.81}
for question 2, and \emph{0.76} for question 3. These scores indicate
substantial or almost perfect agreement among annotators \cite{hallgren2012computing}.
Details on the evaluation interface are in the Appendix.

\begin{algorithm}[h] 
\small 
\caption{Progressive Multi-granularity Decoding} \hspace{\algorithmicindent} 
\textbf{Input}: V: visual embeddings of image I; $E=\{E_c,E_{c-1},\ldots,E_1\}$: series of nested subsequence expressions, $c$: complexity level, $\textrm{max\_length}$: maximum output length
\\ \hspace{\algorithmicindent} \textbf{Output}: $y_c$: generated spatial tokens $y_c$ 
	\begin{algorithmic}[1]     
		\FOR{$i=1$ to $c$}         
			\STATE Initialize $y_i \gets []$         
			\FOR{$j=1$ \textbf{to} $\textrm{max\_length}$} 
				\STATE $y_{i,j} \gets \textrm{\ModelName}(V, E_i, (E_{i-1}, y_{i-1}), y_{i})$             
				\STATE add $y_{i,j}$ to $y_i$         
			\ENDFOR  
		\ENDFOR    
		\STATE $y_c \gets y_i$      
		\STATE \RETURN $y_c$ 
	\end{algorithmic} 
	\label{algo:iterative_decoding_milv} 
\end{algorithm}

\begin{figure}
\includegraphics[width=1\columnwidth]{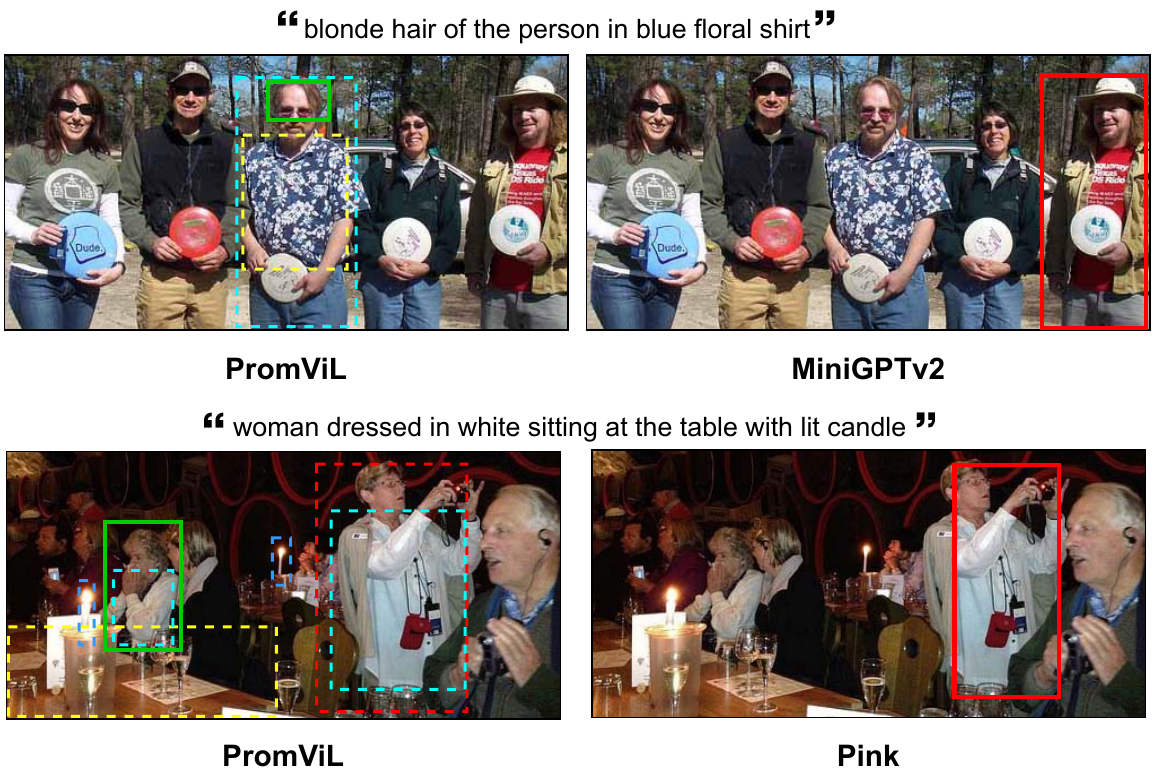}

\caption{\label{fig:Qualitative-figure}Qualitative results on CompoVL-hard.
Solid green: correct boxes, solid red: incorrect. Existing methods
struggle with complex descriptions or multiple similar objects (e.g.,
two " women dressed in white" ). PromViL
leverages simpler expressions (dashed boxes) to accurately locate
complex targets. More qualitative results in Appendix.}
\end{figure}

\subsection{Training and Inference}

\subsubsection{Training}

Our $\ModelName$ framework takes as input an image $I$, represented
by visual embeddings $V$ and a series of nested subsequence expressions
$E=\left\{ E_{1},...,E_{c-1},E_{c}\right\} $, with $E_{c}$ is a
complex compositional expression at level-$c$ complexity ($c>1$).
Multiple expressions can exist at each level of complexity. We maintain
the standard language modeling objective of predicting the next word
token based on the preceding context. However, we explicitly train
$\ModelName$ to visually ground, particularly through utilizing the
nested subsequences in the $\Dataset$ dataset. The model is trained
to generate spatial tokens $y_{i}$ referred to by text expression
$E_{i}$ of level-$i$ complexity in consideration feedback received
from lower levels of complexity. Indicating $<E_{i-1},y_{i-1}>$ as
responses from the previous level of complexity,  our $\ModelName$
framework is trained by optimizing an averaged autoregressive next
token prediction loss across expressions of all levels of complexity:

\[
\text{\ensuremath{\mathcal{L}}}=\frac{1}{\mid E\mid}\sum_{i=1}^{c}\text{log}P(y_{i}|V,E_{i},<E_{i-1},y_{i-1}>);
\]
For data without nested subsequences (e.g., VQA, instruct-follow)
we follow to the common practices employed in other $\TaskName$ such
as LlaVa \cite{liu2024visual} and Kosmos-1 \cite{huang2024language},
where $y_{i}$ are language-based responses to a generic prompt.

In practice, our $\ModelName$ fine-tunes existing $\TaskName$ on
the $\Dataset$ dataset with input representations for language expression
$E_{i}$ of level-$i$ complexity as below: \textit{``\textless s\textgreater\textless img\textgreater V\textless /img\textgreater\textless grounding\textgreater We
can see in the image:\textless p\textgreater\{$E_{i-1}$\}\textless /p\textgreater\{$y_{i-1}$\}.
Based on that, we can locate:\textless p\textgreater\{$E_{i}$\}\textless /p\textgreater\{$y_{i}$\}\textless /s\textgreater ''.}

\subsubsection{Inference}

During inference, we do not have access to nested subsequences of
a given input expression $E_{c}$ of level-$c$ complexity. Therefore,
we use a constituency parser \cite{kitaev-etal-2019-multilingual}
to extract nested subsequences from $E_{c}$ by selecting noun phrase
constituents from leaves toward the root (showed in Fig.~\ref{fig:Our-method}).
We then use a dependency parser to identify the semantic dependancies
between entities within the subsequences, ultimately to remove level-one
expressions that do not satisfy referential entities. This is to
ensure our decomposed nested subsequences to have the same structure
with our data generation process in $\Dataset$. Once these nested
subsequences $E=\{E_{c},E_{c-1},...,E_{1}\}$ are available, we progressively
prompt the model where the generated response to a prior level provides
a clue for the next level. Algorithm 1 describes $\ModelName$'s decoding
process during inference.

\begin{figure}[h]
\begin{centering}
\includegraphics[width=0.85\columnwidth]{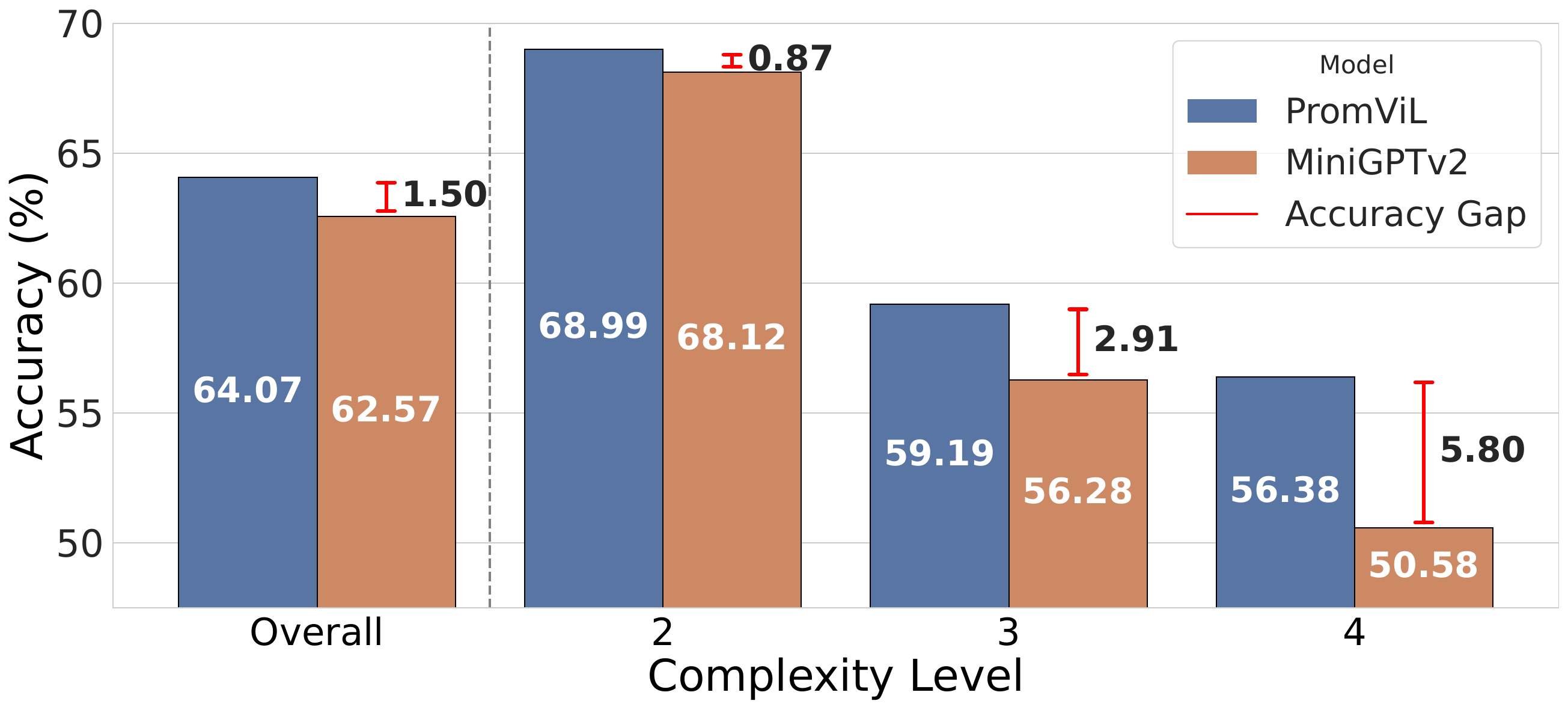}
\par\end{centering}
\caption{\label{fig:Accuracy-comparison-between}Accuracy comparison between
$\protect\ModelName$ and MiniGPTv2 on the $\protect\TestData$ dataset.}
\end{figure}

\begin{table}[t]
\begin{centering}
\begin{tabular*}{1\columnwidth}{@{\extracolsep{\fill}}c>{\centering}p{0.6cm}>{\centering}p{1.6cm}>{\centering}p{1.7cm}}
\toprule 
{\small{}Model} & {\small{}Params} & {\small{}Test Acc. (RefCOCOg)} & \multicolumn{1}{>{\centering}p{2cm}}{{\small{}Acc.}{\small\par}

{\small{}(CompoVL-hard)}}\tabularnewline
\midrule
\midrule 
{\small{}Pink{*}} & {\small{}7B} & {\small{}83.70} & {\small{}62.03}\tabularnewline
\midrule
{\small{}MiniGPTv2{*}} & {\small{}7B} & {\small{}84.66} & {\small{}62.57}\tabularnewline
\midrule
\textbf{\small{}$\ModelName^{\ddagger}$} & {\small{}7B} & \textbf{\small{}85.32} & \textbf{\small{}68.92}\tabularnewline
\midrule
\midrule 
{\small{}Kosmos-2} & {\small{}1.6B} & {\small{}61.65} & {\small{}55.37}\tabularnewline
\midrule 
{\small{}F-Kosmos-2} & {\small{}1.6B} & {\small{}62.69} & {\small{}59.12}\tabularnewline
\midrule 
\textbf{\small{}$\ModelName$} & {\small{}1.6B} & \textbf{\small{}65.28} & \textbf{\small{}64.07}\tabularnewline
\bottomrule
\end{tabular*}
\par\end{centering}
\centering{}\caption{\label{tab:Performance-on-CompoVG_test:} Existing $\protect\TaskName$
excel in understanding simple concepts in RefCOCOg but struggle with
complex expressions in $\protect\TestData$. \emph{F-Kosmos-2:} Kosmos-2
finetuned on the $\protect\Dataset$ excluding nested subsequences.
$\protect\ModelName^{\ddagger}$: our framework finetuned with MiniGPTv2.
({*}) are supervised on region-caption data from VG.}
\end{table}

\begin{table*}
\begin{centering}
\resizebox{0.9\textwidth}{!}{%
\begin{tabular*}{1\textwidth}{@{\extracolsep{\fill}}ccccccccccc}
\toprule 
 & Model & Params & \multicolumn{2}{c}{RefCOCOg} & \multicolumn{3}{c}{RefCOCO+} & \multicolumn{3}{c}{RefCOCO}\tabularnewline
\midrule
\midrule 
 &  &  & val & test & val & testA & testB & val & testA & testB\tabularnewline
\midrule
\multirow{3}{*}{Supervised} & Pink & 7B & 83.70 & 83.70 & 81.40 & \textbf{87.50} & 73.70 & 88.30 & 91.70 & 84.00\tabularnewline
 & MiniGPTv2 & 7B & 84.44 & 84.66 & 79.97 & 85.12 & 74.45 & 88.69 & 91.65 & 85.33\tabularnewline
\cmidrule{2-11} \cmidrule{3-11} \cmidrule{4-11} \cmidrule{5-11} \cmidrule{6-11} \cmidrule{7-11} \cmidrule{8-11} \cmidrule{9-11} \cmidrule{10-11} \cmidrule{11-11} 
 & $\ModelName^{\ddagger}$ & 7B & \textbf{85.91} & \textbf{85.32} & \textbf{81.65} & 87.20 & \textbf{75.62} & \textbf{89.92} & \textbf{92.76} & \textbf{86.77}\tabularnewline
\midrule 
\multirow{5}{*}{Zero-shot} & CoVLM & 1.4B & 60.87 & 61.91 & 47.62 & 50.93 & 44.16 & 48.19 & 53.17 & 43.18\tabularnewline
 & CoVLM & 2.8B & 61.23 & 62.33 & 48.87 & 52.51 & 44.71 & 49.32 & 53.67 & 44.49\tabularnewline
 & Pink & 7B & 59.1 & 60.1 & 43.9 & 50.7 & 35.0 & 54.1 & 61.2 & 44.2\tabularnewline
 & Kosmos-2 & 1.6B & 60.57 & 61.65 & 45.48 & 50.73 & 42.24 & 52.32 & 57.42 & 47.26\tabularnewline
\cmidrule{2-11} \cmidrule{3-11} \cmidrule{4-11} \cmidrule{5-11} \cmidrule{6-11} \cmidrule{7-11} \cmidrule{8-11} \cmidrule{9-11} \cmidrule{10-11} \cmidrule{11-11} 
 & $\ModelName$ & 1.6B & \textbf{64.44} & \textbf{65.28} & \textbf{49.54} & \textbf{53.46} & \textbf{44.92} & \textbf{57.89} & \textbf{62.75} & \textbf{52.89}\tabularnewline
\bottomrule
\end{tabular*}}
\par\end{centering}
\centering{}\caption{\label{tab:Performance-on-Grounding}Comparison on visual grounding
tasks. $\protect\ModelName^{\ddagger}$: our model finetuned with
MiniGPTv2.}
\end{table*}

\begin{table*}
\begin{centering}
\resizebox{0.9\textwidth}{!}{%
\begin{tabular*}{1\textwidth}{@{\extracolsep{\fill}}>{\centering}p{2cm}ccccccc>{\centering}p{1.4cm}>{\centering}p{1.47cm}>{\centering}p{1.69cm}}
\toprule 
\multirow{2}{2cm}{Model} & \multicolumn{7}{c}{GQA val} & \multicolumn{3}{c}{GQA-OOD}\tabularnewline
\cmidrule{2-11} \cmidrule{3-11} \cmidrule{4-11} \cmidrule{5-11} \cmidrule{6-11} \cmidrule{7-11} \cmidrule{8-11} \cmidrule{9-11} \cmidrule{10-11} \cmidrule{11-11} 
 & acc. $\uparrow$ & bin. $\uparrow$ & open $\uparrow$ & consis. $\uparrow$ & valid. $\uparrow$ & plaus. $\uparrow$ & dist. $\downarrow$ & acc\_all $\uparrow$ & acc\_tail $\uparrow$ & acc\_head $\uparrow$\tabularnewline
\midrule 
Kosmos-8K & 40.55 & 43.86 & 37.45 & 64.12 & 74.82 & 71.24 & 13.71 & 32.65 & 31.23 & 33.53\tabularnewline
\midrule 
Kosmos-16K & 41.44 & 44.70 & 38.38 & 66.19 & 75.14 & 71.61 & \textbf{13.33} & 33.83 & 32.36 & 34.74\tabularnewline
\midrule 
$\ModelName$ & \textbf{45.07} & \textbf{52.15} & \textbf{38.44} & \textbf{68.40} & \textbf{83.47} & \textbf{79.80} & 14.49 & \textbf{37.84} & \textbf{33.77} & \textbf{40.33}\tabularnewline
\bottomrule
\end{tabular*}}
\par\end{centering}
\centering{}\caption{\label{tab:Performance-on-Compositional}Performance of $\protect\ModelName$
in comparison with two finetuned versions of Kosmos-2 on the GQA validation
split and the GQA-OOD dataset. Abbreviations for results on GQA: acc.
- accuracy, bin. - binary, consis. - consistency, valid. - validity,
plaus. - plausibility and dist. - distribution. Abbreviations for
results on GQA-OOD: acc\_all - overall accuracy, acc\_tail - out-of-distribution
accuracy, and acc\_head - in-distribution accuracy{\small{}.$\uparrow$:
the higher the better. $\downarrow$: the lower the better.}}
\end{table*}

%% file: experiment.tex
\begin{table}
\begin{centering}
\resizebox{0.95\columnwidth}{!}{%
\begin{tabular*}{1\columnwidth}{@{\extracolsep{\fill}}>{\centering}p{2.5cm}cc}
\toprule 
Model & \multicolumn{1}{c}{Test Acc.} & \multicolumn{1}{c}{Val. Acc.}\tabularnewline
\midrule
\midrule 
Kosmos-8K & 47.24 & 46.82\tabularnewline
\midrule 
Kosmos-16K & 47.96 & 47.74\tabularnewline
\midrule 
$\ModelName$ & \textbf{50.28} & \textbf{50.31}\tabularnewline
\bottomrule
\end{tabular*}}
\par\end{centering}
\centering{}\caption{\label{tab:Performance-on-Visual7W.}Performance of $\protect\ModelName$
in comparison with other finetuned versions of Kosmos-2 on Visual7W.}
\end{table}

\subsection{Implementation Details:}

Unless otherwise stated, our $\ModelName$ is achieved by fine-tuning
Kosmos-2 \cite{peng2024grounding} on the $\Dataset$ dataset, adhering
to its default hyperparameters. We perform LoRA \cite{hu2021lora}
tuning with r=64, learning rate 1e-4, warm-up ratio 0.1, and batch
size 4. LoRA targets all linear layers in the model (85M parameters,
$\sim$4.9\% of Kosmos-2's). Fine-tuning takes around 7 hours on a
single NVIDIA V100 GPU. We use spaCy \cite{spacy} for dependency
parsing and Berkeley Neural Parser \cite{kitaev-etal-2019-multilingual}
for constituency parsing. We evaluate the effectiveness of $\ModelName$
on various visual grounding and downstream tasks, including zero-shot
referring expression grounding tasks and VQA tasks.

\subsection{Experimental Results\label{subsec:Experimental-Results}}

\textbf{$\TaskName$ on $\TestData$ subset:} To provide insights
into the failure of existing $\TaskName$ in compositional visual
grounding, we experiment with various state-of-the-art models, including
Pink \cite{xuan2024pink}, MiniGPTv2 \cite{chen2023minigpt}, Kosmos-2
\cite{peng2024grounding}, on our $\TestData$ subset. We report \emph{top-1
accuracy} of bounding box generation.

Tab.~\ref{tab:Performance-on-CompoVG_test:} and Fig.~\ref{fig:Qualitative-figure}
show these models struggle on this subset both quantitatively and
qualitatively. While these models could understand simple cross-entity
relations (typically two-entity relations) in most situations, such
as those in the RefCOCOg \cite{mao2016generation}, they perform poorly
on more complex compositional expressions in our $\TestData$. For
example, Pink and MiniGPTv2 experience a drastic drop from in accuracy
from over 80.0\% to just over 62.0\%. Smaller models like Kosmos-2
tend to face greater challenges in generalizing their understanding
of primitive concepts to complex relational concepts.

We also finetune these models on our newly established $\Dataset$
dataset to enhance their compositional visual grounding capabilities
(Tab.~\ref{tab:Performance-on-CompoVG_test:}, F-Kosmos-2). Results
indicate that even with exposure to training data containing complex
compositional expressions, the improvement is marginal. Kosmos-2 achieves
only 59.12\% after fine-tuning compared to 55.37\% previously. $\ModelName$
benefits from both having access to complex expressions in $\Dataset$
and advancements in modeling, reaching 64.07\% accuracy.

\begin{figure}[h]
\centering{}\includegraphics[width=0.85\columnwidth]{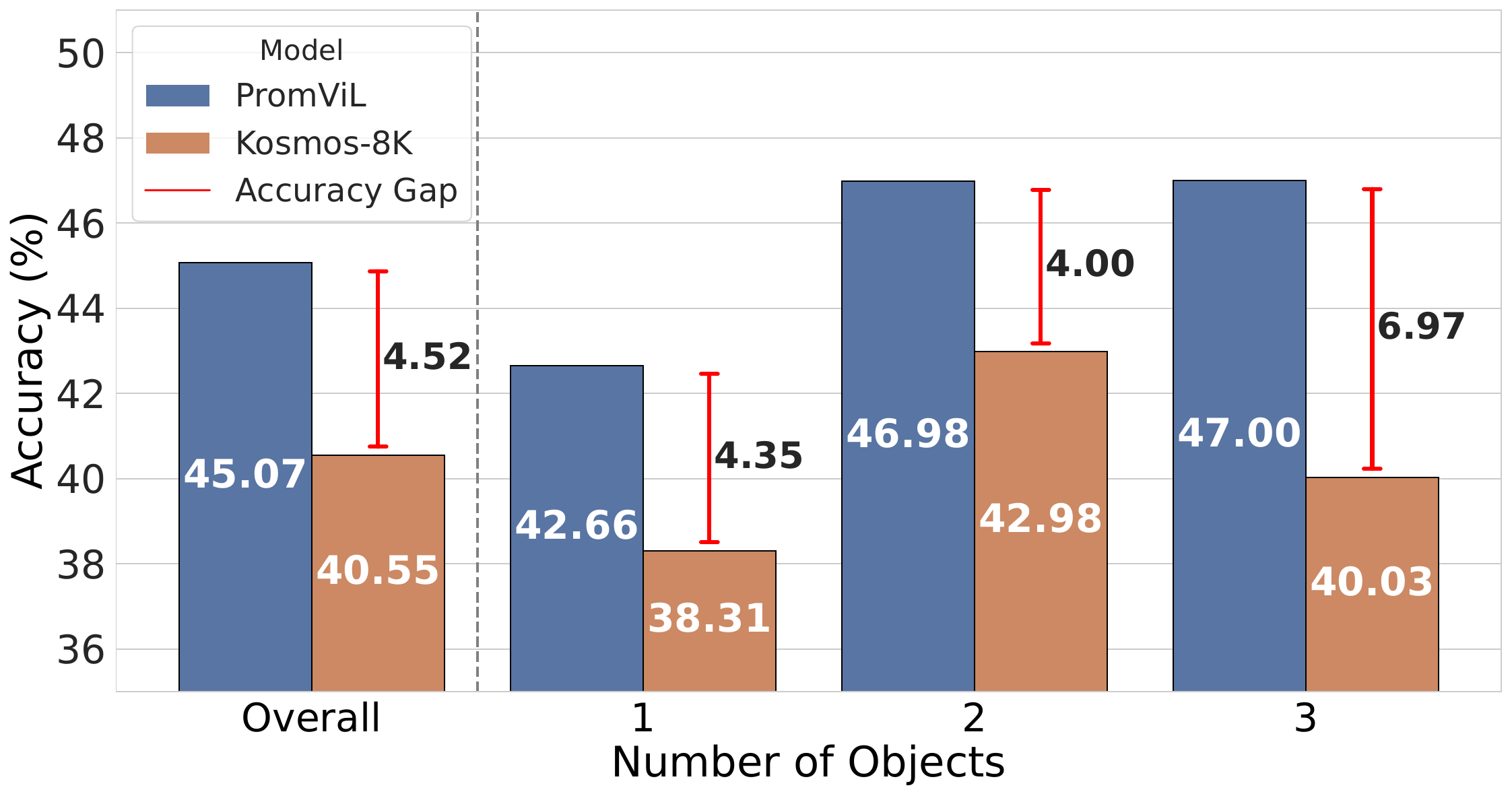}\caption{\label{fig:Accuracy-comparison-between-1}Accuracy comparison between
$\protect\ModelName$ and Kosmos-8K on the GQA\_val dataset.}
\end{figure}

\textbf{Referring expression tasks:} To validate our proposed method
$\ModelName$'s generalization capability on visual grounding task,
we conduct evaluations on well-established benchmarks: RefCOCOg \cite{mao2016generation}
and RefCOCO/RefCOCO+ \cite{kazemzadeh2014referitgame}. For zero-shot
settings, our $\ModelName$ is based on Kosmos-2 while for supervised
settings, we finetune MiniGPT2.

$\ModelName$ consistently outperforms all recent $\TaskName$ across
all datasets in both zero-shot and supervised settings (See Tab. \ref{tab:Performance-on-Grounding}).
In zero-shot settings, it improves upon Kosmos-2 by nearly 4.0 points
and over 5.5 points on average for RefCOCOg and RefCOCO, respectively.
Among these datasets, RefCOCO contains a significantly larger number
of same-type objects \cite{yu2016modeling}, thus benefiting the most
from the $\ModelName$'s multi-granularity V-L alignment. Compared
to other $\TaskName$, $\ModelName$ surpasses all of them with a
considerably smaller model size. Performance gaps are even more pronounced
when comparing $\ModelName$ to other models of comparable sizes.
Similar results are observed on the supervised settings. These results
strongly support the necessity of constructing a \emph{chain of reasoning
cues} to effectively handle complex compositional expressions.

\begin{table}
\begin{centering}
\resizebox{0.9\columnwidth}{!}{%
\begin{tabular*}{1\columnwidth}{@{\extracolsep{\fill}}>{\centering}p{4cm}>{\centering}p{1.5cm}>{\centering}p{0.9cm}>{\centering}p{0.8cm}}
\toprule 
Model & CompoVL

-hard & \multicolumn{2}{>{\centering}p{1.7cm}}{RefCOCOg}\tabularnewline
\midrule
\midrule 
 &  & val & test\tabularnewline
\midrule 
All levels included & \textbf{64.07} & \textbf{64.44} & \textbf{65.28}\tabularnewline
\midrule 
Intermediate levels removed & 61.22 & 63.48 & 63.44\tabularnewline
\midrule 
Only highest level included & 59.12 & 61.25 & 62.69\tabularnewline
\midrule 
Only simplest level included & 53.26 & 59.63 & 59.92\tabularnewline
\bottomrule
\end{tabular*}}
\par\end{centering}
\centering{}\caption{\label{tab:The-effect-of}Ablation studies on the RefCOCOg dataset
and $\protect\TestData$ subset.}
\end{table}

\textbf{Zero-shot compositional VQA tasks:} We also validate $\ModelName$'s
ability to perform compositional VQA tasks on the\textbf{ }GQA \cite{hudson2019gqa}
and GQA-OOD \cite{kervadec2021roses} datasets. Additionally, we challenge
$\ModelName$ with zero-shot grounded VQA tasks using the Visual7W
\cite{zhu2016visual7w} dataset, requiring the model to select the
correct bounding box from four options based on a given question.
Refer to the Appendix for implementation details and prompts for all
models on these tasks. 

We compare against two finetuned versions of the original Kosmos-2:
(1) \emph{Kosmos-8K}: finetuned on the $\Dataset$ dataset excluding
all nested subsequences (2) \emph{Kosmos-16K}: also finetuned on the
$\Dataset$ dataset excluding all nested subsequences, but it is added
with additional 8K VQA pairs randomly sampled from the VQA data of
the VG dataset.

Tab~\ref{tab:Performance-on-Compositional} details experimental
results on the two chosen datasets. As shown, $\ModelName$ outperforms
Kosmos-8K and Kosmos-16K with accuracy gaps of approximately 4.0 points
and 3.5 points, respectively. More importantly, $\ModelName$ is considerably
better than these two baselines in terms of validity and plausibility.
This suggests that clues from lower levels help constrain the model's
search space, ultimately leading to more relevant answers to input
queries. Furthermore, the differences in results of Kosmos-8K and
Kosmos-16K indicate that the increasing size of VQA training data
by twice yields minimal improvements, suggesting inherent limitations
in the modeling of Kosmos.

Regarding the GQA-OOD dataset, $\ModelName$ surpasses both Kosmos-8K
and Kosmos-16K in both out-of-distribution (acc\_tail) and in-distribution
(acc\_head) settings. This indicates that our $\ModelName$ model
is more effective in reducing the reliance on spurious linguistic
correlations between input queries and generated answers.

On zero-shot grounded VQA setting (See Tab.~\ref{tab:Performance-on-Visual7W.}),
we observe that while increasing the size of VQA samples in the training
set does not bring much effect (Row 1 vs. Row 2), the benefits of
using V-L grounding progressively from low-level complexity expressions
to higher-level ones are substantial. This highlights the crucial
role of improved visual grounding capabilities in solving these tasks.

\subsection{Model Analysis and Ablation Studies}

To quantify the the relations between performance improvements and
the complexity levels of input expressions in our $\ModelName$ model,
we analyze its performance on each group of input expressions with
the same level of complexity (See Fig. \ref{fig:Accuracy-comparison-between}).
As shown, the performance gap between our $\ModelName$ and MiniGPTv2
widens as input expressions become more complex. This clearly reaffirms
the benefits of leveraging reasoning cues at lower levels to pave
a reasoning path toward solving more complex tasks.

Similar trends are observed in the GQA dataset evaluation (See Fig.~\ref{fig:Accuracy-comparison-between-1}).
$\ModelName$ consistently improves as scene complexity increases,
achieving 42.66\%, 46.98\%, and 47.0\% accuracy when the number of
objects in visual scenes grows from 1 to 3. In contrast, Kosmos-8K
struggles with increasing object count, dropping from 42.98\% accuracy
with 2 objects to 40.03\% with 3 objects.

Moreover, we conduct an extensive set of ablation studies on $\TestData$
and RefCOCOg to assess the contribution of each model component to
the overall accuracy. Our experiments include: 1)\emph{ All levels
included}: a full model, which features the chain of reasoning at
all levels, ranging from the simple to complex. 2)\emph{ Intermediate
levels removed}: all intermediate-level subsequences are removed.
The model is trained using only expressions of level-one and the highest
level of complexity. 3)\emph{ Only highest level included}: only includes
expressions having the highest level of complexity in a series. 4)\emph{
Only simplest level included}: only includes expressions at the lowest
level of complexity. Tab. \ref{tab:The-effect-of} confirms that
relying on either fine-grained processes and coarse-grained processes
are not sufficient, especially when facing complex compositional queries
as in $\Dataset$.

%% file: discussion.tex
This paper introduced $\ModelName$, a hierarchical framework for
enhancing LVLMs' compositional reasoning capabilities. By progressively
aligning visual and textual information at multiple granularities,
$\ModelName$ enables effective learning of complex compositional
vision-language tasks. We also introduced $\Dataset$, a dataset containing
nested compositional vision-language pairs with varying levels of
complexity, which can be readily used by existing $\TaskName$ to
enable progressive reasoning ability. Experimental results demonstrated
PromViL's superior performance on various benchmarks, advancing the
state-of-the-art in grounded compositional visual reasoning.